\pdfoutput=1

\documentclass[11pt]{article}

\usepackage[preprint]{acl}

\usepackage{times}
\usepackage{latexsym}

\usepackage[T1]{fontenc}

\usepackage[utf8]{inputenc}

\usepackage{microtype}

\usepackage{inconsolata}

\usepackage{makecell}

\usepackage{graphicx}

\usepackage{placeins}

\title{Light-R1: Curriculum SFT, DPO and RL for Long COT from Scratch and Beyond}

\author{%
Liang Wen$^{1}$ \quad Yunke Cai$^{1}$ \quad Fenrui Xiao$^1$ \quad Xin He$^1$ \quad Qi An$^1$ \quad Zhenyu Duan$^1$ \\
\textbf{Yimin Du}$^1$ \quad \textbf{Junchen Liu}$^1$ \quad \textbf{Lifu Tang}$^1$ \quad \textbf{Xiaowei Lv}$^{1,2}$ \\
\textbf{Haosheng Zou}$^1$ \quad \textbf{Yongchao Deng}$^1$ \quad \textbf{Shousheng Jia}$^1$ \quad \textbf{Xiangzheng Zhang}$^1$ \\
$^1$Qiyuan Tech \quad $^2$Renmin University\\
\texttt{zhangxiangzheng@360.cn}
}

\begin{document}

\maketitle

\begin{abstract}

This paper introduces Light-R1, an open-source suite for training long reasoning models using reproducible and cost-effective methodology. Given the proprietary nature of data used in the DeepSeek-R1 series, we develop an alternative approach leveraging exclusively public data and models. Our curriculum training progressively increases data difficulty, combined with multi-staged post-training. Our Light-R1-32B model, trained from Qwen2.5-32B-Instruct, outperforms DeepSeek-R1-Distill-Qwen-32B in math reasoning.
Experimental results show that this curriculum approach becomes more effective when distinct, diverse datasets are available for different training stages: fine-tuning DeepSeek-R1-Distilled models (pre-tuned by DeepSeek team on proprietary data) with 3,000 challenging examples from our curriculum dataset yielded state-of-the-art 7B and 14B models, while the 32B model, Light-R1-32B-DS performed comparably to QwQ-32B and DeepSeek-R1. 
Furthermore, we extend our work by applying GRPO on long reasoning models. Our final Light-R1-14B-DS achieves SOTA performance among 14B models in math, with AIME24 \& 25 scores of 74.0 and 60.2 respectively, surpassing many 32B models and DeepSeek-R1-Distill-Llama-70B. Despite math-focused training, Light-R1-14B-DS demonstrates strong cross-domain generalization.
Light-R1 represents a significant advancement in making sophisticated reasoning models more accessible and implementable in real-world applications. Our models, training data and code have been made available at \url{https://github.com/Qihoo360/Light-R1}.

\end{abstract}


\begin{figure}[ht]
\centering
\includegraphics[width=0.5\textwidth]{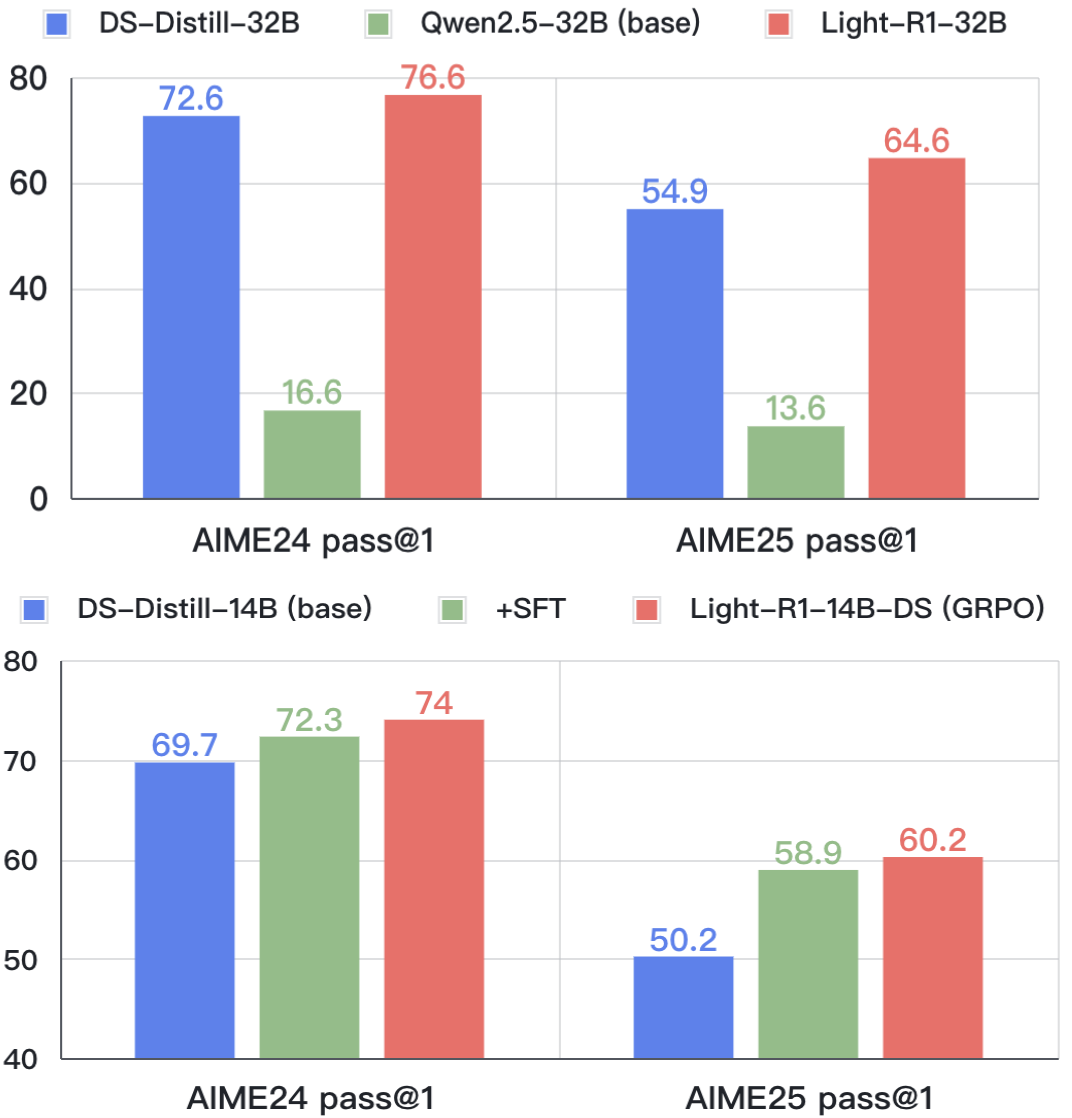
}
\caption{Reproducible state-of-the-art long COT models \textbf{(top)} developed from scratch (=short-COT base), \textbf{(bottom)} derived from DeepSeek-R1-Distill models (=long-COT base), via curriculum learning strategy.}
\label{fig:first}
\end{figure}


\section{Introduction}
\label{sec:intro}
Since the release of DeepSeek-R1 \citep{deepseekai2025deepseekr1incentivizingreasoningcapability}, long chain-of-thought \citep{o1,cot,k1.5,lightman2023letsverifystepstep} reasoning has gained widespread popularity in both foundational AI models and various industrial AI applications. However, deploying full-capacity R1-level models (typically 70B+ parameters, DeepSeek-R1 with 671B parameters) incurs prohibitive computational costs \citep{deepseekai2025deepseekr1incentivizingreasoningcapability,qwq32b}. The resource barrier of training and deploying the giant models makes them impractical for edge devices and real-time applications. This limitation has sparked growing interest in developing compact yet capable models under a few 10B parameters that can perform extended long COT - a critical requirement for mathematical problem solving, algorithmic planning, and scientific analysis.
To address this challenge, we present our work on the Light-R1 series. 

As a foundation for our research, we first established robust and reproducible evaluation protocols that rigorously reproduce the evaluation results reported in \citet{deepseekai2025deepseekr1incentivizingreasoningcapability}. Building upon this reliable framework, our research systematically addresses three fundamental challenges through innovative algorithmic and engineering advancements.

The first challenge involves curating an efficient dataset for Post-Training, a critical factor for long-COT optimization \citep{ye2025limoreasoning,muennighoff2025s1,li2025limrrlscaling}. We collected diverse open-source reasoning data covering mathematical reasoning, logical deduction, and algorithmic problem-solving. After preprocessing to remove duplicates and standardize formatting, we implemented a two-stage difficulty filtering methodology using DeepScaleR-1.5B-Preview \citep{deepscaler2025} and DeepSeek-R1-Distill-Qwen-32B models to quantify difficulty based on pass rates.

The second challenge then emerges as how to optimize the utilization of this dataset. While conventional approaches typically employ a single SFT stage~\citep{deepseekai2025deepseekr1incentivizingreasoningcapability,xu2025redstardoesscalinglongcot,bespoke_stratos,yu2024metamathbootstrapmathematicalquestions}, our preliminary experiments with our 32B model revealed significant limitations—approximately 20\% of training data still exhibited pass rates below 50\% across 10 runs, indicating insufficient knowledge assimilation from heterogeneous difficulty datasets. To address this, we implemented a multi-staged curriculum training strategy comprising two consecutive SFT stages with progressively increasing difficulty, followed by a DPO stage~\citep{rafailov2023direct}. Although recent work has explored different curriculum strategies for long-COT training~\cite{luo2025wizardmathempoweringmathematicalreasoning,min2024imitateexploreselfimprovereproduction,xi2024traininglargelanguagemodels,yuan2025agentrtraininglanguagemodel}, our approach demonstrates superior performance: our Light-R1-32B model, trained from Qwen2.5-32B-Instruct \citep{qwen2.5}, outperforms DeepSeek-R1-Distill-Qwen-32B in mathematical reasoning.

The third challenge arises from implementing the final component of Post-Training — Reinforcement Learning \citep{shaoDeepSeekMathPushingLimits2024,wang2024mathshepherdverifyreinforcellms,ouyang2022training,schulman2017proximalpolicyoptimizationalgorithms,schulman2018highdimensionalcontinuouscontrolusing} — to further enhance model performance. We are excited to report our successful reinforcement learning training of Light-R1-14B-DS. While recent research has shown success in training base models \citep{zeng2025simplerl,OpenReasonerZero2025,liu2025oatzero}, smaller models \citep{zeng2025simplerl,deepscaler2025}, or larger models with intensive computational resources \citep{qwq32b}, our long-COT RL Post-Training represents the first demonstration of simultaneous increases in both response length and reward scores on long-COT 14B models without the initial length reduction typically observed. This breakthrough demonstrates that carefully designed curriculum strategies can overcome the previously documented scalability limitations of RL in smaller models \cite{gao2023scaling}.

The key contributions of this work include:
\begin{itemize}
\item A detailed, fully open-source Curriculum Post-Training approach to train long-COT models from scratch. The multi-stage curriculum training incrementally builds reasoning capacity through difficulty-progressive data exposure, requiring only \$1000 training cost (6 hours on 12×H800 GPUs). This approach is validated on Qwen2.5-32B-Instruct and could be easily migrated to 7B and 14B models.

\item A well established SFT stage 2 dataset of 3k mostly math questions that could significantly improve not only SFT stage 1 but also all DeepSeek-R1-Distill models, resulting in our SOTA 7B model Light-R1-7B-DS.

\item First demonstration of RL effectiveness on 14B models for mathematical reasoning, achieving around 2\% absolute improvement compared with before-RL, resulting in our SOTA 14B model Light-R1-14B-DS.
\end{itemize}



\begin{table}
\centering
\caption{Reproduction of \citet{deepseekai2025deepseekr1incentivizingreasoningcapability} and \citet{qwq32b} evaluation results on AIME24 \citep{AIME2024} pass@1 averaged over 64 runs.}
    \begin{tabular}{ccc}
        \hline
        \textbf{Model} & \textbf{Paper} & \textbf{Ours} \\
        \hline
        DS-distill-32B & 72.6 & 72.3 \\
        DS-distill-14B & 69.7 & 69.3 \\
        DS-distill-7B & 55.5 & 54.0 \\
        QWQ-32B & 79.5 & 78.5 \\
        \hline
    \end{tabular}
\label{tab:eval}
\end{table}

\section{The Origin of Everything: Stable and Trustworthy Evaluation of Long-COT Models}

Following \citet{deepseekai2025deepseekr1incentivizingreasoningcapability}, long-COT models are commonly deployed with sampling temperature 0.6. While long-COT models generally perform better with sampling than with greedy decoding, it brings more burden for model evaluation as multiple samples for each question may be required, contrary to previous viable approaches of greedy decoding for evaluation \citep{song2024good}.

\citet{deepseekai2025deepseekr1incentivizingreasoningcapability} generates 64 responses per query to estimate pass@1. We have verified this choice, witnessing large deviation of over 3 points using 16 responses or fewer across different runs of the same model. Such randomness is unacceptable to compare model performances.

For stable and trustworthy evaluation, we adapted \cite{deepscaler2025}'s evaluation code for all our evaluation runs.
Our evaluation code and logs are all released.

We can reproduce all DeepSeek-R1-Distill models' and QwQ's scores as reported in \citet{deepseekai2025deepseekr1incentivizingreasoningcapability,qwq32b} as shown in Tab. \ref{tab:eval} with 64 samples per query, with deviation around 1 point.

\section{Light-R1-32B: Long-COT from Scratch with Curriculum SFT \& DPO}

While numerous studies \citep{ye2025limoreasoning,muennighoff2025s1,openthoughts,openr1} have open-sourced efforts to replicate DeepSeek-R1 using models of various sizes, ranging from 1.5B to 32B, none has reached similar performance on the challenging mathematics competitions AIME24 \& 25, where DeepSeek-R1-Distill-Qwen-32B scored at 72.6 \& 54.9.

\begin{figure*}[ht]
\centering
\includegraphics[width=0.9\textwidth]{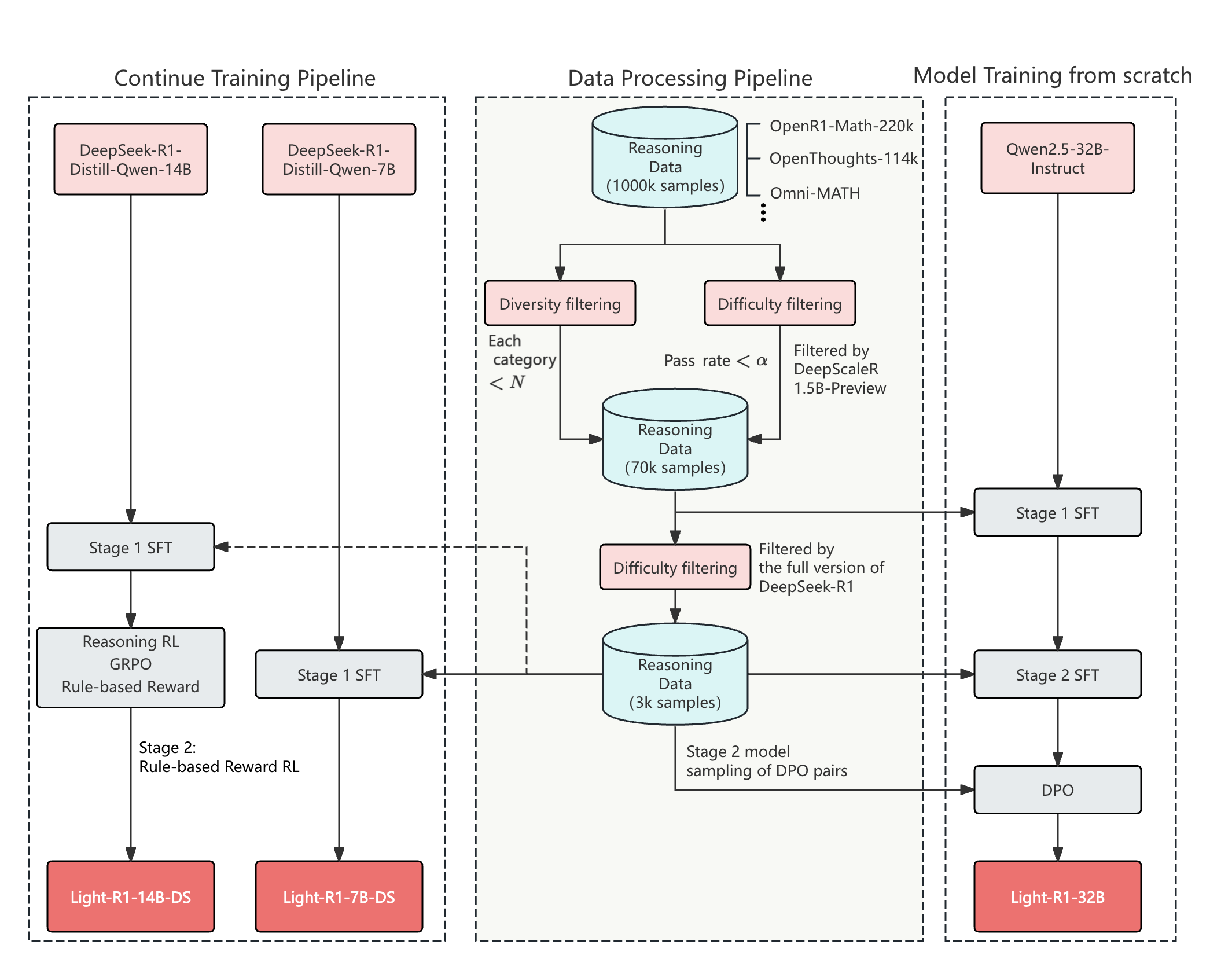}
\caption{Overview of training pipeline of Light-R1 series.}
\label{fig:overview}
\end{figure*}

We present our data processing and Post-Training pipeline in this section as illustrated by Fig. \ref{fig:overview}.

\subsection{Data Preparation}
The whole data preparation process spans data collection, data decontamination and data generation, detailed as follows.

\subsubsection{Data Collection}
We began by collecting various sources of math questions with groundtruth answers.
Iterating over all possible sources by the time, we collected around 1000k math questions as the seed set. See Appendix~\ref{sec:appendix_b} for more details about the data sources.


All data are aggregated together to form around 1000k math questions as the seed set. 
Within this 1000k data, we kept only math questions with groundtruth answers. Questions without groundtruth answers could be used as synthetic data by letting multiple strong LLMs vote for groundtruths but we left it for future work. 

The data is then filtered for diversity, where we tagged each question with an in-house tagging system and downsample categories with excessive data.

\subsubsection{Data Decontamination}



We evaluated data contamination in several open-sourced datasets. Our analysis revealed that MATH-500 \citep{hendrycks2021measuringmathematicalproblemsolving} contains tens of compromised questions that are either identical or differ only in numerical values. AIME 24 and 25 remain uncontaminated, though caution is needed when incorporating AIME data through 2023. Further details are provided in Appendix~\ref{sec:appendix_a}.

Light-R1 underwent comprehensive decontamination using exact matching (excluding digits to filter questions with only numerical changes) and N-gram (N=32) matching against AIME24\&25, MATH-500, and GPQA \citep{rein2023gpqagraduatelevelgoogleproofqa}.

\subsubsection{Data Generation}
\label{sec:datagen}
With a diverse and clean dataset, we generate comprehensive chain-of-thought (COT) responses for supervised fine-tuning (SFT). However, not all data points are equally valuable for training, and distilling DeepSeek-R1 can be resource-intensive whether through API queries or local deployment. We therefore implemented difficulty-based filtering on the dataset to retain only sufficiently challenging questions, inspired by recent advances in training long reasoning models \citep{deepscaler2025,ye2025limoreasoning,muennighoff2025s1}.

We initially employ \citet{deepscaler2025}'s DeepScaleR-1.5B-Preview model to generate responses for each question, as this model offers a good balance of efficiency and capability. Only questions with a pass rate $<\alpha$ were selected for DeepSeek-R1 queries, resulting in approximately 76k data points. After obtaining DeepSeek-R1 responses, we retained only questions with correct long-COT answers. For questions with multiple correct responses, we randomly selected one long-COT answer for SFT. Through this process, we constructed an SFT dataset exceeding 70k examples, featuring prompts filtered for both diversity and difficulty, with long-COT responses generated by DeepSeek-R1 and validated against ground truth.

However, direct training on this dataset alone did not yield satisfactory results regardless of the number of training epochs. Upon analyzing the trained model's performance across different question types, we discovered the need for additional training on more challenging problems.
Consequently, we implemented a second stage of difficulty filtering using the full version of DeepSeek-R1 instead of DeepScaleR-1.5B-Preview. This stage retained only questions with pass rate $<\alpha$ and questions where DeepSeek-R1's sampled responses were neither uniformly correct nor uniformly incorrect, resulting in a Stage 2 SFT dataset of approximately 3k examples.
Notably, this refined dataset demonstrated such high quality that training exclusively on it produced performance improvements across all DeepSeek-R1-Distill models, as we will discuss in Section \ref{sec:stage2data}.

\subsection{Curriculum Post-Training}
Our approach consists of three stages, detailed hyperparameters can be found in Appendix \ref{sec:appendix_c}.:
\begin{enumerate}
    \item \textbf{SFT Stage 1}: Training on 76k filtered mathematical problems
    \item \textbf{SFT Stage 2}: Fine-tuning on 3k most challenging problems
    \item \textbf{DPO Optimization}: Preference-based optimization using verified response pairs
\end{enumerate}

SFT stages are trained with the curriculum data strategy as discussed in Sec. \ref{sec:datagen}. For DPO, we implemented a semi-on-policy approach using the NCA loss \citep{chen2024noise}. Rejected responses were sampled from our SFT-stage-2 model with verified incorrect answers. Since some rejected responses reached lengths of 32k tokens or more, we utilized the DPO implementation with sequence parallelism from 360-LLaMA-Factory \citep{360-llama-factory}. For chosen responses, we used verified correct answers from DeepSeek-R1. While we had previously employed fully on-policy DPO extensively, we discovered that for challenging mathematical problems, using chosen responses from significantly stronger models yielded better results.

\subsection{Results}

We observe consistent improvements across our curriculum SFT \& DPO post-training stages (Tab. \ref{tab:stageimprove}). Following DPO, we use the TIES-merging \citep{yadav2023tiesmergingresolvinginterferencemerging} method from the \citet{goddard-etal-2024-arcees} toolkit to merged models from SFT-stage2, DPO, and another DPO variant (AIME24 score: 74.7) that had special tokens inadvertently removed from rejected responses, the resulting merged model demonstrates additional performance gains. Although our mathematics-focused training led to some forgetting on untrained GPQA scientific questions, Light-R1-32B still demonstrates strong generalization capabilities.


\begin{table*}[!htbp]
\setlength{\tabcolsep}{4pt}
\centering
\begin{minipage}{0.48\textwidth}
\fontsize{9pt}{13pt}\selectfont
\centering
\begin{tabular}{lcccc}
\hline
\textbf{Stage} & \textbf{AIME24} & \textbf{AIME25} & \textbf{GPQA} & \textbf{LCB} \\
\hline
Instruct (base) & 16.6 & 13.6 & 48.8 & 24.6 \\
\texttt{ +}SFT-stage1 & 69.0 & 57.4 & 64.3 & 42.9 \\
\texttt{ +}SFT-stage2 & 73.0 & 64.3 & 60.6 & 42.0 \\
\texttt{ +}DPO & 75.8 & 63.4 & 61.8 & N/A \\
\texttt{ +}Model Merging & 76.6 & 64.6 & 61.8 & 44.7 \\
Light-R1-32B & \textbf{76.6} & \textbf{64.6} & 61.8 & 44.7 \\
\hline
\end{tabular}
\caption{Stage-wise performance improvement of our Light-R1-32B. We observe a decrease in GPQA (Science QA) scores beginning from STF-stage2, indicating a partial degradation of the model's generalization capabilities during extensive math-focused training. However, Light-R1-32B still demonstrates strong generalization 
compared to the base model. }
\label{tab:stageimprove}
\end{minipage}
\hfill
\begin{minipage}{0.48\textwidth}
\fontsize{9pt}{13pt}\selectfont
\centering
\begin{tabular}{lcccc}
\hline
\textbf{Model} & \textbf{AIME24} & \textbf{AIME25} & \textbf{GPQA} & \textbf{LCB} \\
\hline
DS-distill-7B & 55.5 & 39.2 & 49.1 & 34.6 \\
Light-R1-7B-DS & \textbf{59.1} & \textbf{44.3} & 49.4 & 38.4 \\
\hline
DS-distill-14B & 69.7 & 50.2 & 59.1 & 52.9 \\
Light-R1-14B-DS' & \textbf{72.3} & \textbf{58.9} & 60.3 & 55.9 \\
\hline
DS-distill-32B & 72.6 & 54.9 & 62.1 & 58.8 \\
Light-R1-32B-DS & \textbf{78.1} & \textbf{65.9} & \textbf{68.0} & 66.1 \\
\hline
\end{tabular}
\caption{Effectiveness of the 3k data from SFT stage2. Fine-tuning on stronger base models, which presumably utilize datasets orthogonal to ours, consistently enhances performance across all model sizes. The notation \textbf{Light-R1-14B-DS'} refers to the SFT-only version of our final Light-R1-14B-DS model, which subsequently undergoes an additional stage of GRPO RL training.}
\label{tab:allimprove}
\end{minipage}
\end{table*}

\FloatBarrier

\subsection{High-Quality Data is All You Need}
\label{sec:stage2data}
Considering DeepSeek-R1-Distill-Qwen models as a stronger version of our SFT stage 1, we performed SFT stage 2 with the 3k stage 2 data on top of DeepSeek-R1-Distill-Qwen models.

Surprisingly as Tab. \ref{tab:allimprove}, we could achieve universal improvement on DeepSeek-R1-Distill-Qwen models with this 3k data alone, demonstrating the high quality of the stage 2 data. It may also be because this 3k data is to some extent orthogonal to DeepSeek-R1-Distill-Qwen models' 800k SFT data, hence such easy improvement.

GPQA performance is unexpectedly high for Light-R1-32B-DS, despite the absence of domain-specific training in science and code domains, suggesting that stronger base models may benefit from stronger generalization capacities. In contrast, Light-R1-7B-DS, while trained on identical data curriculum, exhibits improvements confined solely to in-domain tasks.


\section{Light-R1-14B-DS: Reinforcement Learning from Long-COT Models}
\label{sec:rl}



\begin{table*}[t]
\setlength{\tabcolsep}{4pt}
\centering
\begin{minipage}{0.48\textwidth}
\fontsize{11pt}{15pt}\selectfont
\includegraphics[width=\textwidth]{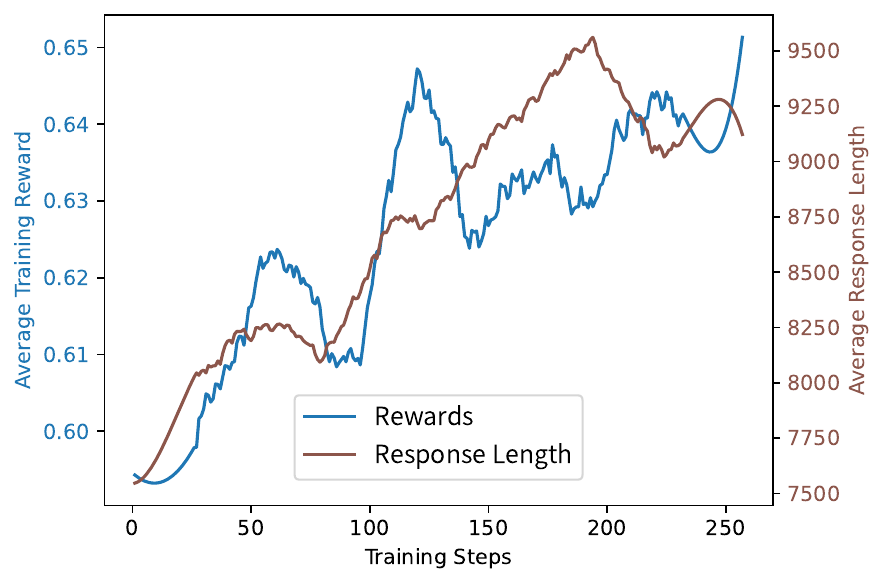}
\captionof{figure}{RL Learning curves of response length and train-reward, smoothed with Savitzky-Golay filter.}
\label{fig:rl}
\end{minipage}
\hfill
\begin{minipage}{0.48\textwidth}
\fontsize{9pt}{13pt}\selectfont
\centering
\begin{tabular}{lcccc}
\hline
\textbf{Model} & \textbf{AIME24} & \textbf{AIME25} & \textbf{GPQA}
& \textbf{LCB} \\
\hline
DS-distill-14B & 69.7 & 50.2 & 59.1 & 52.9 \\
\texttt{ + }SFT & 72.3 & 58.9 & 60.3 & 55.9 \\
\texttt{ + }GRPO epoch1 & 72.3 & 57.8 & N/A & 56.6 \\ 
\texttt{ + }GRPO epoch2 & 73.4 & 60.5 & N/A & 56.5 \\
\makecell[l]{Light-R1-14B-DS \\ (GRPO epoch3)} & \textbf{74.0} & \textbf{60.2} & \textbf{61.7} & 56.0 \\
GRPO dataset-v2 & 75.0 & 65.0 & 62.6 & 57.9 \\ 

\hline
\end{tabular}
\caption{RL performance improvement of Light-R1-14B-DS. Notably, we observe out-of-domain improvement in GPQA, indicating that reinforcement learning on mathematics-focused datasets potentially facilitates generalization across diverse domains.}
\label{tab:rl}
\end{minipage}
\end{table*}



We conduct our reinforcement learning experiments on DeepSeek-R1-Distill-Qwen-14B. To the best of our knowledge, this is the first publicly documented work demonstrating significant improvement in performance through RL on already long-COT 14B models. 

Previous studies by \citet{deepseekai2025deepseekr1incentivizingreasoningcapability}, \citet{yuanWhatsPPOsCollapse2025}, and \citet{zhang2025dpor1} have shown that smaller models (with 32 billion parameters or fewer) can reach high performance levels through distillation from larger reasoning models. However, 
further improvement via RL (Reinforcement Learning) on already long-COT finetuned models is not yet widely reached by the community and is not as easily reachable as \emph{zero} RL (Sec. \ref{sec:intro}). 
While \citet{deepscaler2025} successfully demonstrated promising RL training on a smaller model DeepSeek-R1-Distill-Qwen-1.5B, we encountered challenges in replicating similar results with the larger DeepSeek-R1-Distill-Qwen-14B model using the same recipe.

After weeks of investigation, we arrived at 
our final RL solution consisting of a two-pass process, drawing inspiration from our effective curriculum SFT attempt and \citet{cuiProcessReinforcementImplicit2025}. The process is as follows:
\begin{enumerate}
    \item \textbf{Offline Data Selection}: Use Light-R1-7B-DS to sample results of RL training prompts. Keep only the prompts whose pass rate is between 0.25 and 0.625.
    \item \textbf{Online Reinforcement Learning}: Apply GRPO on the filtered dataset.
\end{enumerate}

In our observation, offline data selection plays a critical role. It filters out prompts that are too easy or too hard and ensures that the training data aligns with our rule-based answer verifier. When manually checking data with a pass rate of 0, we found that over half of the prompt answers are either unverifiable (due to containing text or complex conditional expressions) or incorrect. We utilize Light-R1-7B-DS as the difficulty estimation model because it is more efficient and demonstrates similar performance to larger models in terms of pass@64. Additionally, we use a model verifier to re-check data with a pass rate of 0. By filtering out the mis-verified data, we can successfully identify difficult prompts for future curriculum reinforcement learning.

We choose GRPO \citep{shaoDeepSeekMathPushingLimits2024} as the optimization algorithm and implement it based on verl \citep{sheng2024hybridflow}. We also employ two techniques to stabilize the RL training process: modified version of length reward \citep{yeoDemystifyingLongChainofThought2025} with weaker preference for short correct answers and importance sampling weight clipping \citep{minimaxMiniMax01ScalingFoundation2025}. 

For length control, we adopt a modified version of the approach proposed by \cite{yeoDemystifyingLongChainofThought2025}. Specifically, we clip the shortening reward when answers are correct to prevent initial length collapse. This technique helps maintain a reasonable answer length during training, ensuring that the model does not overly shorten its responses at the beginning of the learning process.

Regarding importance sampling weight clipping, we implement a broader two-sided clipping mechanism. Our observations have shown that occasional large positive policy ratios combined with negative advantages can lead to loss spikes, disrupting policy optimization. This two-sided clipping technique was also implemented in our previous experiments, in parallel with the findings reported by \citet{minimaxMiniMax01ScalingFoundation2025}. By clipping the importance sampling weights, we can limit the influence of extreme values and make the training process more stable.

We use a rule-based reward and the de-duplicated version of the Big-Math dataset (\citet{albalak2025bigmathlargescalehighqualitymath}). The experiments are conducted on a cluster of 16 * 8 A100 GPUs. The offline data selection process takes 4 hours, while the online reinforcement learning takes 26 hours to complete 140 steps and 42 hours to complete 220 steps.

As can be seen from Fig. \ref{fig:rl}, our RL training demonstrates expected behavior: simultaneous increase in response length and reward score. No interesting length dropping in the beginning.
We evaluated RL epochs 1 and 2 after we finished training 3 epochs. As shown in Tab. \ref{tab:rl}, although first two epochs seem to bring not much improvement, the healthy RL training curves offer us confidence to continue training.
Light-R1-14B-DS is finally RL trained for around 3 epochs, or 220 steps.

\section{Conclusion}

Our Light-R1 series addresses the challenge of training long reasoning models under resource constraints. We successfully train a long-COT model from scratch through our curriculum training strategy. Our carefully curated 3K dataset demonstrates remarkable transferability across various model sizes, significantly enhancing DeepSeek-R1-Distill models and establishing new performance benchmarks for models with 7B, 14B, and 32B parameters. Additionally, we investigate the efficacy of reinforcement learning when applied to a strong multi-stage finetuned base model, achieving superior performance while maintaining stable response length growth throughout the training process.

These advancements not only democratize access to R1-level reasoning capabilities but also provide valuable insights into curriculum design, data efficiency, and RL scalability for long reasoning models. Our open-source models, datasets, and code aim to accelerate research in developing compact yet powerful reasoning systems, particularly for resource-constrained applications. Future work will explore the integration of enhanced generalization capabilities for long reasoning models and further optimization of RL training efficiency.

\bibliography{latex/main}

\newpage
\appendix
\onecolumn

\section{Light-R1 Series of Models}

\begin{table}[ht]
\centering
\label{tab:alllight}
\caption{Light-R1 models. ``-DS'' = from DeepSeek-R1-Distill, otherwise from Qwen-Instruct.}
\begin{tabular}{lccccl}
\hline
\textbf{Model} & \textbf{AIME24} & \textbf{AIME25} & \textbf{GPQA} & \textbf{LCB} & \textbf{Training Recipe} \\
\hline
Light-R1-32B & 76.6 & 64.6 & 61.8    & 44.7  & SFT stage1\&2 + DPO \\
Light-R1-7B-DS & 59.1 & 44.3 & 49.4  & tbd   & SFT stage2 \\
Light-R1-14B-DS & 74.0 & 60.2 & 61.7 & 56.0 & SFT stage2 + GRPO \\
Light-R1-32B-DS & 78.1 & 65.9 & 68.0 & 66.1  & SFT stage2 \\
\hline
\end{tabular}
\end{table}

\section{Dataset composition for full 59K questions}
\label{sec:appendix_b}

\begin{table*}[htbp]
\centering
\caption{\textbf{Composition of the released data.}
Here we summarize the data composition after the first stage diversity and difficulty filtering. Different sources may contain overlapping examples, we use OpenR1-Math-220k as our initial seed dataset, which explains why this source contributes the largest portion of our data.}
\begin{tabular}{p{0.3\linewidth} p{0.4\linewidth} p{0.1\linewidth} p{0.1\linewidth} p{0.1\linewidth}}
\hline
Source & Description & \#Samples \\
\hline
OpenR1-Math-220k~\citep{openr1} & Math problems with two to four reasoning traces generated by DeepSeek R1 for problems from NuminaMath 1.5. & 58224\\
\hline
OpenThoughts-114k~\citep{openthoughts} & Open synthetic reasoning dataset with 114k high-quality examples covering math, science, code, and puzzles & 14214 \\
\hline
OpenMathInstruct-2~\citep{toshniwal2024openmath2} & Math instruction tuning dataset generated using the Llama3.1-405B-Instruct model by Nvidea & 1786 \\
\hline
OmniMath~\citep{gao2024omnimathuniversalolympiadlevel} & Math problems from competitions & 567 \\
\hline
s1K-1.1~\citep{muennighoff2025s1} & Diverse, high-quality \& difficult questions with distilled reasoning traces \& solutions from DeepSeek-R1 & 346 \\
\hline
LIMO~\citep{ye2025limoreasoning} & Three-stage filtered data from the LIMO paper & 246 \\
\hline
hendrycks-math~\citep{hendrycksmath2021} & 12,500 challenging competition mathematics problems. Each problem in MATH has a full step-by-step solution which can be used to teach models to generate answer derivations and explanation & 179
\\
\hline
Ours & In-house math dataset & 3877 \\
\hline
\textbf{Total} & Composite of the above datasets with reasoning traces and solutions & 79439 \\
\hline
\end{tabular}
\label{tab:ds}
\end{table*}

\FloatBarrier


\section{Data Decontamination}
\label{sec:appendix_a}

\begin{table*}[htbp]
    \centering
    \caption{Number of matched prompts in open-source datasets against benchmarks.}
    \begin{tabular}{cccc}
        \hline
        \textbf{Dataset} & \textbf{AIME24+25} & \textbf{MATH-500} & \textbf{GPQA Diamond} \\
        \hline
        \hline
        OpenThoughts-114k & 0 & 100 & 0 \\
        \hline
        Open-R1-Math-220k & 0 & 10 & 0 \\
        \hline
        DeepScaleR-Preview-Dataset & 0 & 196 & 0 \\
        \hline
        LIMO & 0 & 0 & 0 \\
        \hline
        Bespoke-Stratos-17k & 0 & 125 & 0 \\
        \hline
        Open-Reasoner-Zero & 0 & 325 & 0 \\
        \hline
        simplescaling/data\_ablation\_full59K & 0 & 244 & 1 \\
        \hline
        simplescaling/s1K-1.1 & 0 & 3 & 1 \\
        \hline
        ours & 0 & 0 & 0 \\
        \hline
    \end{tabular}
    \label{tab:data-contam}
\end{table*}

\FloatBarrier

\section{Training hyperparameters for Light-R1 series}
\label{sec:appendix_c}

\begin{table*}[htbp]
    \centering
    \caption{Training hyperparameters for Light-R1 series. Sequence length is determined by training data characteristics, except for GRPO where it balances multiple factors: minimizing roll-out computational costs, reducing inference cut-off ratio, and optimizing 32k context evaluation performance. To overcome the limitation of GPU memory for training DPO with 32k context length, we utilize the DPO implementation with sequence parallelism from 360-LLaMA-Factory \citep{360-llama-factory}. Models with the "-DS" suffix derive from the DeepSeek-R1-Distill-Qwen series, while others from Qwen2.5-32B-Instruct.}
    \begin{tabular}{lccc}
        \hline
        \textbf{Model Names} & \textbf{Learning Rate} & \textbf{Batch Size} & \textbf{Seq Length} \\
        \hline
        \hline
        Light-R1-32B SFT Stage1 & $5.0 \times 10^{-5}$ & 96 & 20k \\
        Light-R1-32B SFT Stage2 & $1.0 \times 10^{-5}$ & 32 & 20k \\
        Light-R1-32B DPO & $5.0 \times 10^{-7}$ & 16 & 32k \\
        \hline
        Light-R1-7B-DS & $5.0 \times 10^{-6}$ & 32 & 20k \\
        \hline
        Light-R1-14B-DS-SFT & $5.0 \times 10^{-6}$ & 32 & 20k \\
        Light-R1-14B-DS (GRPO) & $1.0 \times 10^{-6}$ & 128 & 24k \\
        \hline
        Light-R1-32B-DS & $5.0 \times 10^{-6}$ & 32 & 20k \\
        \hline
    \end{tabular}
    \label{tab:hyper}
\end{table*}

\FloatBarrier

\end{document}